\documentclass[runningheads]{llncs}
\usepackage{graphicx}

\usepackage{tcolorbox}
\usepackage{listings}
\usepackage{algorithm}
\usepackage{algpseudocode}
\usepackage{listings}
\newtcolorbox[list inside=prompt,auto counter,number within=section]{prompt}[1][]{
    colbacktitle=black!60,
    coltitle=white,
    fontupper=\footnotesize,
    boxsep=5pt,
    left=0pt,
    right=0pt,
    top=0pt,
    bottom=0pt,
    boxrule=1pt,
    title={#1},
    #1, 
}
\begin{document}
\title{From Large to Tiny: Distilling and Refining Mathematical Expertise for Math Word Problems with Weakly Supervision}

\author{Qingwen Lin\inst{1} \and
Boyan Xu\inst{2} \and
Zhengting Huang\inst{3}\\
Ruichu Cai\inst{1}\thanks{Corresponding author}}

\institute{Dmir Lab, Guangdong University of Technology, Guangzhou, China
\and
\email{qingwen\_lin@foxmail.com}\inst{1}, 
\email{hpakyim@gmail.com}\inst{2}, \email{huang@gmail.com}\inst{3},\email{cairuichu@gmail.com}\inst{1*}}

\maketitle              
\begin{abstract}
Addressing the challenge of high annotation costs in solving Math Word Problems (MWPs) through full supervision with intermediate equations, recent works have proposed weakly supervised task settings that rely solely on the final answer as a supervised signal. Existing leading approaches typically employ various search techniques to infer intermediate equations, but cannot ensure their semantic consistency with natural language descriptions. The rise of Large Language Models (LLMs) like ChatGPT has opened up new possibilities for addressing MWPs directly. However, the computational demands of LLMs make them less than ideal for use in settings where resources are tight. 
In light of these challenges, we introduce an innovative two-stage framework that adeptly transfers mathematical Expertise from large to tiny language models. In \emph{Distillation Stage}, we propose a series of extraction processes that satisfy the properties of MWPs to distill mathematical knowledge from LLMs to construct problem-equation pairs required for supervised training. In \emph{Refinement Stage},
Due to Knowledge distilling method cannot guarantee the full utilization of all data, we further utilize the unsuccessfully searched data effectively by Knowledge Refine method. Finally, We train a small model using distilled data generated through two-stage methods. As our method fully leverages the semantic understanding capabilities during the searching 'problem-equation' pair, it demonstrates significantly improved performance on the Math23K and Weak12K datasets compared to existing small model methods, while maintaining a much lower computational cost than ChatGPT.

\end{abstract}

\section{Introduction}

\begin{figure}[t] 
\includegraphics[width=\textwidth]{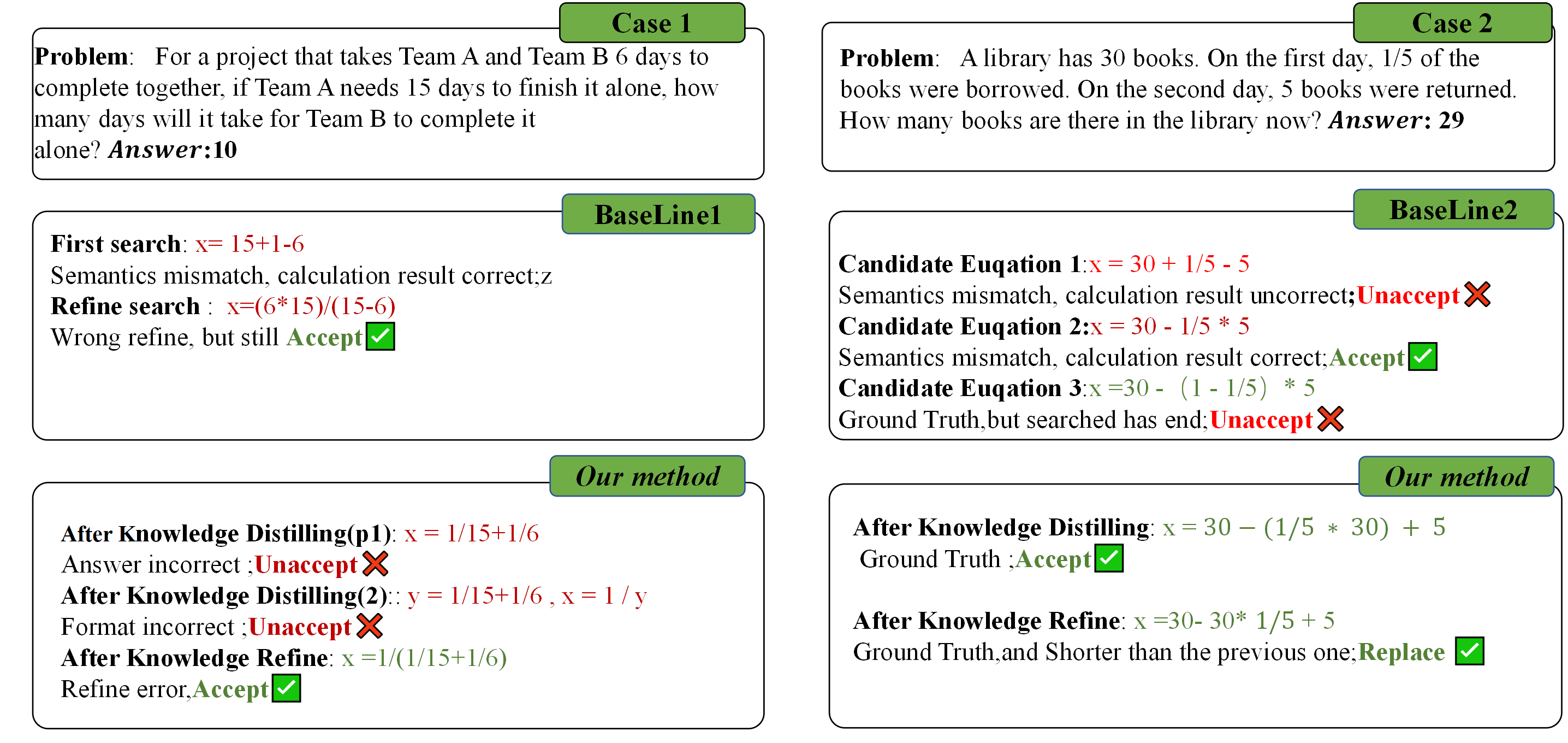}
\caption{Process of Format Correction}\label{fig:case}
\end{figure}

Solving math word problems (MWPs)  is a complex task that provides solutions to mathematical problems written in natural language.  Seq2Seq models have become a common choice for modeling and solving this complex task. The Seq2Seq model trains the MWPs solver by using the 'problem-equation' pair as supervised signals. For instance, encoder-decoder architectures \cite{xie2019goal,wang2017deep,wang2019template,liu2019tree,xiao2023recursive,wu2023graph,shen2020solving}, data augmentation and standardization techniques \cite{liu2020reverse,wang2018translating}, as well as leveraging pre-trained models \cite{liang2021mwp,tan2022investigating,shen2021generate}
. However, current MWPs solvers are primarily trained under fully supervised settings.  It requires annotated 'problem-equation' pairs, which is expensive and time-consuming.

In recent years, researchers have found that the cost of obtaining only the problem and answer is much lower than annotating 'problem-equation' pairs\cite{hong2021learning}. Therefore, the weakly supervised setting is proposed to automatically obtain equations through problems and answers, thus reducing annotation costs. Current weakly supervised methods primarily involve automated rule-based searches for 'problem-equation' pairs, such as random walks \cite{hong2021learning}, beam search \cite{chatterjee2021weakly}, and combinatorial search \cite{liu2022comsearch}. However, these methods may lead to incorrect semantic matching. 
As shown in Figure \ref{fig:case}, the existing method may produce answers consistent with the ground truth calculation, but it fails to match the semantics of the problem, thereby misleading the model's understanding ability. This is because when searching for equations, the semantic matching with the problem is overlooked. We address this data noise as the false-matching problem. On the other hand, Large Language Models (LLMs) like ChatGPT \cite{ouyang2022training}, introduce another method to address annotation costs. ChatGPT leveraging its powerful semantic understanding, can effectively address MWPs without training. However, the parameter size of ChatGPT is significantly larger than existing weakly supervised methods, limiting its implementation with lower computational resources.

Therefore, this paper aims to utilize the semantic understanding capabilities of ChatGPT to train a small model under  weakly supervised setting. This addresses the issue of MWPs solvers false-matching problems in existing weakly supervised methods, especially in environments with limited computational resources. To achieve this goal, we propose \textbf{From Large to Tiny: Distilling and Refining Mathematical Expertise for Math Word Problems with Weakly Supervision (FLTT)}. This method consists of two key steps: \textbf{Knowledge Distilling}, \textbf{Knowledge Refine}.

Firstly, the \textbf{Knowledge Distilling} method leverages the semantic understanding capabilities of ChatGPT. Through an iterative process of auto-generation and correction, high-quality 'problem-equation' pairs that align with the requirements of  small model are generated. As shown in Figure \ref{fig:case},, due to the uncertainty in ChatGPT's outputs, this method may not fully extract all 'problem-equation' pairs from weakly supervised data, leading to the generation of unsuccessfully searched data. To address this, we introduce the \textbf{Knowledge Refine}  method. This method involves finetuning a middle model using successfully searched data and then utilizing this model for iterative searches on unsuccessfully searched data. This process aims to enhance the efficiency of utilizing weakly supervised data, thereby improving the performance of the small model.

Finally, we train a new small model with the distilled data obtained from the \textbf{Knowledge Distilling and Knowledge Refine} methods. As the model progressively shrinks during this process, our approach is termed \textbf{From Large to Tiny (FLTT)}. Experimental results show that this method outperforms existing methods on the Math23K and Weak12K datasets while maintaining much lower computational costs than ChatGPT.

In summary, the contributions of this paper are as follows:

1. We propose a novel weakly supervised method that leverages ChatGPT to assist in searching for 'problem-equation' pairs equations, effectively addressing the issue of erroneous matches present in current methods.

2. We propose a two-stage distillation method based on knowledge distillation and knowledge refinement. This method effectively automatically searches high-quality ‘problem-equation’ pairs for train by small models.

3. The experimental results on the Math23K and Weak12K datasets indicate that the small model trained using our method outperforms all currently available weakly supervised methods with the same parameter quantity. Moreover, in scenarios with an abundance of weakly supervised data, our small model exhibits superior performance compared to zero-shot LLM.

\section{Question Definition}
 The MWPs solver employs a seq2seq model, thus requiring a set of question $X = (x_1, x_2, . . . , x_m)$ and equation $Y = (y_1, y_2, . . . , y_n)$ as the supervised signals. Corresponding answers $A$ are used to estimate the quality of the model. These are shown in Table \ref{tab:math_word_problem}.In fully supervised setting, the dataset includes $X$,$Y$, and $A$. However, in weakly supervised setting, the dataset only includes $X$ and $A$. Therefore, it is necessary to employ specific methods to acquire "question-equation" pairs as the supervising signal of the MWPs solver.

 \section{Methodology}

\subsection{Overview of FLTT}

\begin{figure}[t] 
\includegraphics[width=\textwidth]{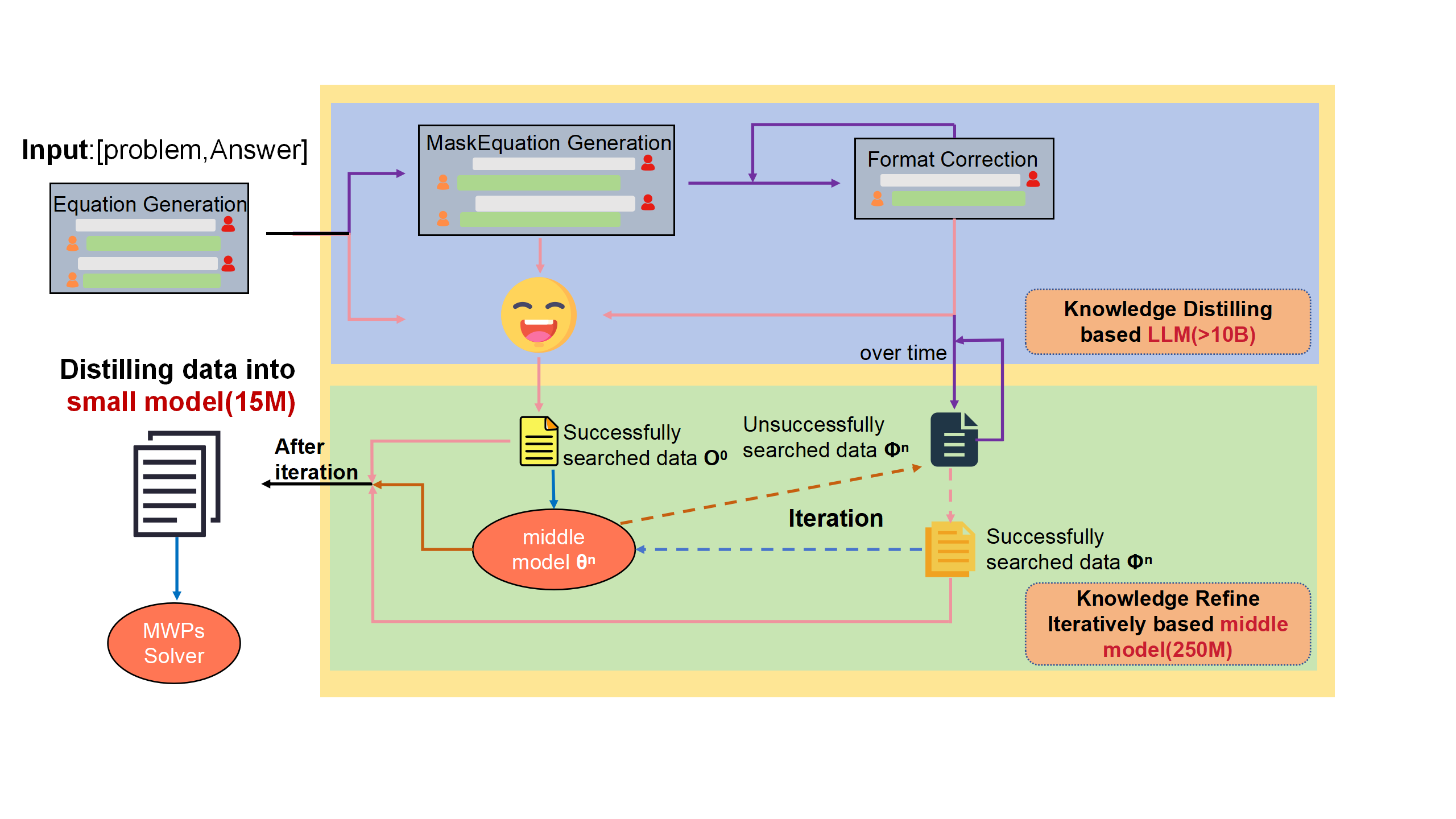}
\caption{A conceptual demonstration of our FLTT method that endows local small models with ChatGPT.}\label{fig:all}
\end{figure}

 We introduce a method called \textbf{From Large to Tiny: Distilling and Refining
Mathematical Expertise for Math Word
Problems(FLTT)} to extract knowledge from LLM to solve math word problems (MWPs). As shown in figure \ref{fig:all}, the FLTT method consists of the following two key steps:

\textbf{(1) Knowledge Distilling}:
As shown in Figure \ref{fig:all}, our goal at this stage is to preliminarily extract knowledge from weakly supervised data using LLM.  Specifically, LLM performs automatic knowledge extraction and error correction to improve data quality and format consistency. However, due to the uncertainty in the outputs of the LLM, this step might generate unsuccessfully searched data.

\textbf{(2) Knowledge Refine}: In this step, our objective is to further extract knowledge from the unsuccessfully searched data and obtain the final distilled data used for training the MWPs solver. we use successfully searched data to finetune a middle model. Then, we try to re-match the unsuccessfully searched data and refine others by the middle model. Compared to the results obtained in the previous step, the distilled data acquired in this step contains knowledge that is more repetitive, refined, and succinct.

Finally, we utilize the distilled data as the supervisory signal for training the final small model.

\subsection{Data processing}\label{processing}

To ensure that the MWPs solver accurately learns knowledge, we need to perform certain processing on all "problem-equation" pairs. Following the existing data processing methods for MWP tasks, we need to transform {[Problem, Equation]} into {[Mask Problem, Mask Equation]}\cite{dong2022survey}. 

Specifically, assuming there are $k$ numbers in the problem $X$, represented as $V = (v_1, v_2, . . . , v_k)$, corresponding to these $k$ numbers, there are $MASK = (mask_1, mask_2, . . . , mask_k)$. In $X$, each $v_i$ is replaced with $v_i [blank] mask_i$. In $Y$, each $v_i$ is replaced with $mask_i$.$y_j$.The results after MASK processing will be shown in Table \ref{tab:math_word_problem}.

\begin{table}[htbp]
\caption{Case of Data}
\centering

\begin{tabular}
{|p{0.9\linewidth}|p{0.9\linewidth}|}
    \hline
    \textbf{Problem:} Andy has 12 apples, Bob has 20 apples, and Bob gives 2 apples to Andy, how many more apples does Bob have than Andy now? \\
    \hline
    \textbf{Equation:} $x = 20 - 12 - 2 - 2$ \\
    \hline
    \textbf{Answer:} 4 \\
    \hline
    \textbf{Mask Problem:} Andy has 12[blank][Mask1] apples, Bob has 20[blank][Mask2] apples, and Bob gives 2[blank][Mask3] apples to Andy, how many more apples does Bob have than Andy now? \\
    \hline
    \textbf{Mask Equation:} $x = [Mask2] - [Mask1] - [Mask3] - [Mask3]$ \\
    \hline

\end{tabular}

\label{tab:math_word_problem}
\end{table}
 
	\subsection{Knowledge Distilling based LLM}

    This paper employs the multi-turn conversational format of ChatGPT as the LLM. In this step, we need to extract preliminary solutions to mathematical problems from ChatGPT.However, due to ChatGPT being a general language model, its output sequence $Y=(y_1, y_2, . . . , y_n)$ often resembles natural language and may not always conform to the  (\ref{eq:SET}). Without annotated "problem-equation" pairs as examples, the commonly used context learning methodes \cite{dong2022survey} are challenging to apply. 
    
    To address this challenge, this paper introduces a \textbf{Knowledge Distilling} method, leveraging multi-turn interactive prompts to encourage ChatGPT for self-correction. This method consists of three stages: \textbf{Equation generation}, \textbf{Masked equation generation}, and \textbf{Format correction}.

Considering that not every search attempt will successfully return results matched (\ref{eq:SET}). In this situation, the small model is unable to learn the corresponding knowledge. Hence, it is necessary to try new searches after each unsuccessful attempt. At the end of each stage, it is essential to perform both format and answer checks on the equations returned by ChatGPT.

\textbf{Format check}: For the results $H_i^k$ returned for the i-th round of the k-th iteration, the processing method in Section \ref{processing} should be applied. It is necessary to ensure that each $y_i \subset H_i^k$ can only be an element from the MASK set, operator set, and constant set for the MWPs solver to accurately learn the knowledge. This is denoted as Equation (\ref{eq:SET}).

\begin{equation}\label{eq:SET}
    {y_j} \subset MASK \cup \{+, -, \cdot, /, \hat{} \} \cup \{1, 100, \pi\}
\end{equation}

\textbf{Result check}: After passing the format check, the results $H_i^k$ returned for the i-th round of the k-th iteration should satisfy the (\ref{eq:answer}).

\begin{equation}\label{eq:answer}
     | H_{i}^{k} - A | < 10^{-4}
\end{equation}

If the "check format" passes successfully, then check the result; otherwise, proceed to the next step. If the result passes the check, accept this equation; otherwise, consider the current search as unsuccessful. If Format check and  Result check pass, we make $Y = H_i^k$ and make $[X, H_i^k]$ as a "problem-equation" pair. We set a maximum of three search attempts;  if $k > 3$, the search is considered unsuccessful.

    \subsubsection{Equation Generation for Problem}

    In this step, considering the robust zero-shot capabilities of ChatGPT, our method aims to make ChatGPT autonomously extract correct problem-solving knowledge by COT prompt and ExtractPrompt.

    The overall process is illustrated in Prompt \ref{prompt:Equation Generation}. First, we guide ChatGPT through a dialogue, prompting it to engage in problem-solving through the interchange of questions X and answers A by COT\cite{wei2022chain} prompt, Then, we aim to extract mathematical knowledge from this process that a smaller model can learn to match (\ref{eq:SET}). ChatGPT generates the equation based on the previous dialogue and [Extract knowledge prompt]·
\begin{prompt}[title={Prompt \thetcbcounter: Equation Generation}, label=prompt:Equation Generation]

User: Provide $[Question]$ and $[Answer]$, and use COT to outline the detailed solution process.  \\
LLM: Return [The process of solving the question]\\
\\
User: Send [Extract Knowledge Prompt]\\
LLM: Return [Equation]\\
\\
Check Format and Check result\\

\end{prompt}

    \subsubsection{MaskEquation Generation for Mask Problem}
Due to ChatGPT's possible to consolidate intermediate steps during the equation generation process. For example, ChatGPT might output "$x=-(12+2+2)$" as "$x=20-16$"., this would lead to inconsistency with the defined (\ref{eq:SET}). This would result in the small model being unable to learn the mathematical knowledge embedded in it.

\begin{prompt}[title={Prompt \thetcbcounter: MaskEquation Generation}, label=prompt:MaskEquation Generation]
The context of "Equation Generation"\\

User: Provide $[Mask Question]$, and use  COT to outline the detailed solution process.  \\
LLM: Return [The process of solving the Mask question.]\\
\\
User: Send [Extract Knowledge Prompt]\\
LLM: Return [Mask Equation]\\
\\
Check Format and Check result\\

\end{prompt}

To address this issue, we employ a method similar to that outlined in \ref{processing} to process the Question into the corresponding MaskQuestion.Guide ChatGPT to generate the corresponding expression while disregarding the numerical values.

This step is similar to the previous one, be illustrated in Prompt \ref{prompt:MaskEquation Generation}. First, ChatGPT combines preceding information and COT to contemplate the Mask Problem. Then, use the [Extract knowledge prompt] to extract the [MASK Equation]. 

\subsubsection{Format Correction for Equation}
Through the preceding steps, it is not guaranteed that all equations will match (\ref{eq:SET}) because the natural language output from ChatGPT exhibits randomness. Therefore, further adjustments to the summarized results are necessary. To address this issue, we have identified five possible scenarios, as shown in Table \ref{tab:Format Correction}.

When one of these erroneous formats is detected, the model is prompted to make corrections. As shown in Table \ref{tab:Format Correction}, if the format remains incorrect after the first correction attempt, we will repeat the correction process up to five times. If the format check still fails after five correction attempts, it is considered a timeout and requires a fresh attempt. 

It is worth noting that ChatGPT's outputs conform to natural language conventions, and, as such, the potential errors it may produce are similarly difficult to enumerate, just as they are with humans. The five scenarios outlined in this paper cannot cover all possible format errors. Therefore, it is inevitable that some unsuccessfully searched data that is denoted as  $[{\omega}]$ will be generated. Correspondingly, there are also successfully searched data, denoted as $[{\mathrm O}]$.
    \begin{table}
    
    \centering
    \caption{Error Type and cases to be checked for "Format Correction"}
    \resizebox{1\textwidth}{!}{
    \begin{tabular}{|c|c|}
        \hline
        \textbf{Error Type}  & \textbf{case}   \\
        \hline
        Improper use of \% & x= temp\_a * temp\_c * temp\_b\%  \\
        \hline
        Expressions using LaTeX notation & x = /frac((/text(temp\_a) /cdot /text(temp\_c) /cdot \\
        \hline
        Multiple expressions appearing & x= temp\_a * y,y=temp\_c* temp\_b/100 \\
        \hline
        Occurrence of text or otherformats & The expression of the problem is (temp\_a * temp\_c * temp\_b) / 100 yuan  \\
        \hline
        Occurrence of non-compliantnumbers and variables & x= (50000 * temp\_c * temp\_b) / 100  \\
        \hline
        Ground Truth & x= (temp\_a  * temp\_c * temp\_b) \\
        \hline
    \end{tabular}%
    }
    
    \label{tab:Format Correction}
\end{table}

\subsection{Knowledge Refine based middle model}

To fully leverage the data $[{\mathrm O}]$, existing methods primarily involve using $[{\mathrm O}]$ as seed data to fine-tune LLM to ensure the desired output data format \cite{gan2023usa}. However, finetuning LLM implies significant resource consumption.

Therefore, the \textbf{Knowledge Refine }  method involves fine-tuning a middle model with data $[{\mathrm O}]$, the middle model denoted as $[{\theta}]$. The parameter size of the middle model remains relatively small compared to the ChatGPT, allowing for a substantial reduction in resource consumption.

Firstly, the middle model $[{\theta}]$ is finetuning using the data $[{\omega}]$ to obtain the finetuned model $[{\theta}^{0}]$. Simultaneously, an empty candidate dataset $[\Phi]$ is initialized.

Next, the model $[{\theta}^{0}]$ is used to perform a Beam-Search to search for the top 5 optimal potential equations. Subsequently, from these 5 results, the one that passes the Result check and is the shortest is selected, as shorter correct results are considered less likely to have redundancy, such as '1-1'.For the found correct shortest answers, they are added to the candidate dataset $[\Phi]$ and removed from the dataset $[{\mathrm O}]$.

Following this, an iterative process is initiated. In the i-th iteration, the model $[{\theta}^{i}]$ is fine-tuned using the dataset $[{\mathrm O}]$ to obtain the refine model $[{\theta}^{i+1}]$. Similar to the previous steps, the model $[{\theta}^{i+1}]$ is used to search for the corresponding equation for problems in the dataset $[{\mathrm O}]$. Any successfully searched data is added to the dataset $[\Phi]$ and removed from $[{\mathrm O}]$. The entire process is summarized as Algorithm \ref{algo:refine_equation}.

\begin{algorithm}
\caption{Knowledge Refine Algorithm}
\label{algo:refine_equation}
\begin{algorithmic}

\State {Initialize dataset $\omega$,${\mathrm O}$,${\Phi}$}
\State \textit{${\theta}^{0}$} $\gets$ \textsc{FineTuning}($\theta $,$\omega$)

\For{\textit{iteration} in 0 to $n$}
    \For{\textit{problem} in \textit{${\mathrm O}$}}
        \State Candidates $\gets$ \textsc{Search}(problem,${\theta}^{i}$)
        \State RightEqs $\gets$ right equation in Candidates
        \If{RightEq != empty}
            \State equation $\gets$ ShortestEqs(RightEq)
            \State \textit{${\Phi}$}.append([problem, \textit{equation}])
            \State \textit{${\mathrm O}$}.pop(problem)
        \EndIf
    \EndFor
    
    \State \textit{${\theta}^{i+1}$} $\gets$ \textsc{FineTuning}(${\theta}^{i}$,${\Phi}$)
\EndFor
\end{algorithmic}
\end{algorithm}

After completing the training, we utilized the model $[{\theta}^{'}]$ and employed the same inference process as previously described to infer data $[{\omega}]$. Choosing passes the Result check and the shorter of the inference results from ChatGPT and the middle model as the final answer, yielding distilled data $[{\omega}^{`}]$. As shown in Figure \ref{fig:case}, it will make data $[{\omega}^{`}]$ more concise than data $[{\omega}]$.

\subsection{Distilling knowledge  to MWPs solver}

This paper employs  Seq2Seq models as MWPs solver, typically, two models are utilized: UniLM \cite{dong2019unified} and T5 model \cite{raffel2020exploring}. this paper simultaneously employs them as the foundational model and employs Distilled data $[{\omega}^{`}]\cup [\Phi]$ as the supervised signals.

Due to the distillation of knowledge from both ChatGPT and the middle model in the dataset $[{\omega}^{`}]\cup [\Phi]$ through two stages, with the model size reducing at each step, hence the method proposed in this paper is referred to as \textbf{From Large to Tiny (FLTT)}.
        
\section{Experimental Results}
    \subsection{Dataset}
     We evaluate our weak supervision method on the Math23K dataset \cite{wang2017deep} and Weak12K \cite{lianggeneralizing}. Math23K contains 23,161 math word problems annotated with solution expressions and answers. It is currently the most widely used dataset for MWPs.Weak12K is a Chinese weakly supervised dataset that contains  12,117  MWPs.Weak12K differs from Math23K in that it was initially created in a weak supervision setting, lacking reference full-supervised equations. Furthermore, the Weak12K dataset is known for its higher difficulty level, which means that many models that perform well on Math23K struggle with this dataset.
     
    \subsection{Baselines}
    
    \textbf{LBF}\cite{hong2021learning}:   LBF  is the first weakly supervised solution for MWPs. It searches for candidate equations through a random walk.
    
    \textbf{ComSearch}\cite{liu2022comsearch}:Comsearch reduces the potential search space of 'problem-equation' pairs through combinatorial methods, thereby improving the recall of weakly supervised data.

     \textbf{WDA}\cite{lianggeneralizing}: WDA enhances the generalization of mathematical problem solvers through the use of a solution buffer and a solution discriminator. 
     
    The above methods use GTS(15M)\cite{xie2019goal} as MWPs solver under a weekly supervised setting. We followed the relevant settings from the original paper and conducted training on our setting about dataset.In addition to the aforementioned small models, we tested the performance of ChatGPT with COT under zero-shot conditions.

     \subsection{Experimental setup}
     This paper employs  RoFormer-Small(15M) \cite{su2021roformer}  with unilm and T5-small(95M)\cite{dong2019unified} as the MWPs solver. Furthermore, for the Knowledge Refine method, we employed the T5-Base(250M) as the middle model. It is still considered a small model in comparison to ChatGPT.  
     
     Other hyperparameters, epochs: 150 for most, 40 for Knowledge Refine; dropout: 0.25; optimizer: Adam; learning rates: 1e-5 for Knowledge Refine/RoFormer, 1e-4 for T5; evaluation metric: accuracy.

\begin{table}[htbp]
    
    \centering
     \caption{Result of Comparison Experiments}
    \resizebox{0.7\textwidth}{!}{%
        \begin{tabular}{|c|c|c|c|}
            \hline
            \textbf{Model} & \textbf{Math23K} & \textbf{Weak12K} & Parameter Size\\
            \hline
            \textbf{Full Supervision} & & & \\
            \hline
            GTS & 75.6 & -& 15M \\
            roformer+unilm  & 76.1 & -& 15M \\
            \hline
            \textbf{Small model} & \textbf{Math23K} & \textbf{Weak12K} & Parameter Size\\
            \hline
            LBF & 55.4 & 19.6& 15M \\
            ComSearch  & 60.5 & -& 15M \\
            WDA & 55.1  & 33.9& 15M \\
            \hline
            \textbf{LLM} & \textbf{Math23K} & \textbf{Weak12K} & Parameter Size\\
            \hline
            ChatGPT & 69.4 & \textbf{60.5}& $>$10B \\
            \hline
            \textbf{Ours} & \textbf{Math23K} & \textbf{Weak12K} & Parameter Size\\
            \hline
            FLTT-roformer  & \textbf{65.1} & \textbf{41.2}& 15M\\
            FLTT-T5& \textbf{71.8} & 46.1& 95M \\
            \hline
        \end{tabular}
    }
   
    \label{MainResult}
\end{table}

    \subsection{Comparative Experiments Result}
\
    
    \subsubsection{Experimental Results on Small Models}
     At an equivalent parameter count, our method achieved a 4.6\% and 7.3\% performance improvement on the Math23K and weak12k. The MWPS solver in this paper differs from existing weakly supervised methods. To eliminate this difference, we also conducted comparative experiments on Math23K under fully supervised setting. Experimental results indicate that there is little difference in performance between the two MWPS solvers.
     
     Therefore, this paper represents the state-of-the-art (SOTA) in solving MWPs with small models under weakly supervised settings.

    \subsubsection{Experimental Analysis on Small Models}

    The reason our method outperforms existing methods is because the quality of the data we search is higher. As shown in Table \ref{tab:Recall}, our method does not exhibit an advantage in recalling "problem-equation" pairs in problem texts with different numbers of variables compared to the LBF and ComSearch methods. However, despite not having a clear advantage in data recall, our method shows significant performance improvements. This is evidently because the data quality searched by our method is significantly better than Baseline.

    The higher data quality achieved by our method is primarily due to the consideration of the semantic information in the problem text during both the  ChatGPT and middle model. In contrast, the Baseline method mainly focuses on the matching between the answers and the recalled equations during the equation generation process, lacking consideration of the semantic information in the problem text. Therefore, plenty of low-quality synthetic data generated by existing methods may mislead the model, thereby performance inferior to our method.

\begin{table*}[t]
    
    \centering
    \caption{Recall of  Comparison Experiments}
        \resizebox{0.4\textwidth}{!}{%
        \begin{tabular}{|c|c|c|c|}
                \hline
                \textbf{VAR*}  & \textbf{LBF} & \textbf{ComSearch} & \textbf{FLTT} \\
                \hline
                1 & 91.5 & 67.0 & 52.5 \\
                2 & 86.8 & 93.4 & 89.9 \\
                3 & 88.8 & 96.4 & 92.7 \\
                4 & 31.1 & 98.1 & 85.4 \\
                5 & 25.5 & 94.4 & 77.5 \\
                $>$6 & 38.4 & 73.8 & 60.1 \\
                \hline
            \end{tabular}
    }
    \label{tab:Recall}
\end{table*}

\subsubsection{Experimental Results and Analysis on LLM}
The experiments indicate that small models trained using our method outperform ChatGPT under specific conditions. As shown in Table \ref{tab:Recall}, after transitioning from a  RoFormer(15M) to T5(95M), there is a 2.4\% performance improvement on Math23K compared to zero-shot ChatGPT using COT.

However, on the Weak12K dataset, the performance of our model is noticeably weaker than ChatGPT. This is because the Weak12K dataset has a smaller amount of data compared to Math23K. Therefore, we believe that when there is sufficient weakly supervised data, our method outperforms zero-shot ChatGPT. Fortunately, weakly supervised data is remarkably easy to acquire in large quantities compared to fully supervised data. We can consider that our method presented can achieve mathematical capabilities comparable to ChatGPT at a lower data cost and with minimal computational resource requirements.

    \subsection{Ablation Experiments}

In this chapter, we need to delve into the roles played by different steps in our method.

In Table \ref{tab:AblationRecall}, I conduct an importance analysis by showcasing detailed recall rates at different stages of our method. In Table \ref{tab:AblationResult}, we analyze the performance of the test set by examining the role of different modules in the Knowledge Refine method.
\begin{table*}[t]
    
    \centering
    \caption{RECALL of Ablation Experiments}
    \resizebox{0.7\textwidth}{!}{%
    \begin{tabular}{|c|c|c|}
        \hline
        \textbf{Knowledge Distilling} & \textbf{Math23K} & \textbf{Weak12K} \\
        \hline
        Equation Generation & 31.7& 23.1\\
        MaskEquation Generation  & 27.8& 26.4\\
        Format Correction &6.8&6.3\\
        Successfully searched&66.5&55.8\\
        \hline
        \textbf{Knowledge Refine} & \textbf{Math23K}&\textbf{Weak12K}\\
        \hline
        SelfFLTT  &85.1 &76.6\\
        FLTT  & \textbf{88.4}&\textbf{76.6}\\
        \hline
    \end{tabular}%
    }
    \label{tab:AblationRecall}
\end{table*}
 In Table \ref{tab:AblationRecall} and Table \ref{tab:AblationResult}, the "SelfFLTT" and "FLTT" represent using RoFormer(15M) and T5(95M) as the middle model.  In Table \ref{tab:AblationResult}, the "finetuning model" and "refine model" represent the models $[\theta^{0}]$ and $[\theta^{'}]$ from Knowledge Refine and the middle model is RoFormer(15M).

    \begin{table*}[t]
    
    \centering
    \caption{Result of Ablation Experiments}
    \resizebox{0.4\textwidth}{!}{%
    \begin{tabular}{|c|c|c|}
        \hline
        \textbf{Model}  & \textbf{Math23K} & \textbf{Weak12K} \\
        \hline
        finetuning & 55.5 & 35.3 \\
        refine & 58.7  & 37.0 \\
        SelfFLTT  & 62.2  & 39.0 \\
        FLTT*  & 63.5 & 39.7 \\
        FLTT  & \textbf{65.1} & \textbf{41.2} \\
        
        \hline
    \end{tabular}%
    }
    
    \label{tab:AblationResult}
\end{table*}  

\subsubsection{Ablation Experiments of Knowledge Distilling}

The experimental results indicate that the steps of equation generation and MaskEquation generation play a primary role in the Knowledge Distillation method. It achieves recall rates of 59.5\% and 49.5\%, accounting for 89.4\% and 88.7\% of the total recall for the Knowledge Distillation method. The recall rates for these two steps are roughly the same., indicating that these play equally crucial and decisive roles in this method.

The role of Format Correction is relatively minor, it could only successfully recall 6.8\% and 6.3\% data. The main reason is the limited enumeration of error types, inevitably resulting in format errors that cannot be corrected. Additionally, the uncertainty in the output of ChatGPT itself makes it can't ensure understanding and successfully correct errors. Therefore, the Knowledge-Refine method is needed for further correction.

    \subsubsection{Ablation Experiments of Knowledge Refine}

In Table \ref{tab:AblationResult}, we observe a performance decrease of 2.9\% and 2.2\% when switching the middle model from T5-base (250M) to RoFormer-small (15M). This is because larger models have stronger semantic understanding capabilities, thereby generating more semantically matched high-quality data. Correspondingly, the quality and quantity of data determine the performance of the MWPs solver.

This viewpoint is validated in Table \ref{tab:AblationRecall}. In  Weak12K, RoFormer-small (15M) and T5-base (250M) have comparable recall rates, but the final results are worse. That indicates the lower quality of data generated by RoFormer contributes to this outcome. In Math23K, the recall rate of T5 is higher than that of RoFormer-small, resulting in a higher improvement magnitude. Therefore, we should choose a middle model with a larger parameter count to ensure the quality and recall rate of the data. However, considering the increased computational resource consumption associated with larger models, we opt for the relatively smaller T5 Base model (250M parameters) as the middle model in the Knowledge Refine method.

In Table \ref{tab:AblationResult}, the Refine model generated during the Knowledge Refine training process exhibits a notable improvement of 3.2\% and 1.7\% compared to the finetuning model. This indicates that the data obtained through the Knowledge Refine method is of high quality, enabling continuous incremental training to enhance the model's performance. However, the performance of the Refine model is significantly weaker than the selfFLTT model, indicating that the iterative search approach in the Knowledge Refine method can affect the generalization of the Refine model. Therefore, it is necessary to retrain an unaffected initial model using distilled data.

In the final step of the \textbf{Knowledge Refine} method, we utilized the model $[{\theta}^{'}]$ and employed the same inference data $[{\omega}]$. We found that 41.9\% and 28.6\% of the data respectively became more concise. To explore whether this setup could improve data quality, we conducted an ablation experiment where we did not re-infer data $[{\omega}]$, denoted as FLTT\*. Experimental results showed that this setup improved model performance by 1.6\% and 1.5\%. Therefore, our hypothesis was correct.

\section{Conclusion}
 This paper presents a novel weakly-supervised method called FLTT. By distilling mathematical knowledge through multiple stages, the model continuously reduces in size, thereby outperforming existing methods on the Math23K and Weak12K datasets. The FLTT method distills knowledge through two stages: Knowledge Distillation and Knowledge Refinement, effectively utilizing data generated by the LLM and further optimizing the performance of the middle model. Experimental results demonstrate that FLTT can achieve mathematical problem-solving capabilities comparable to those of large language models at a lower data cost and with minimal computational resource requirements. This work showcases the potential of multi-stage knowledge distillation in enhancing the capabilities of small models in resource-constrained environments.
 
\section{Related Work}

Methods for addressing Math Word Problems (MWPs) have predominantly relied on deep learning techniques, particularly encoder-decoder architectures. These approaches encompass tree-based models \cite{xie2019goal,wang2019template,liu2019tree,xiao2023recursive}, data augmentation strategies \cite{liu2020reverse}, pretrained model utilization \cite{liang2021mwp,tan2022investigating,shen2021generate}, analogical reasoning frameworks \cite{liang2022analogical}, and optimization-based methodologies \cite{shen2021generate}. However, these methodologies typically operate under fully supervised conditions, incurring significant annotation costs.

To mitigate annotation expenses, weakly supervised learning has emerged as a promising alternative for MWPs. Notable approaches include employing random walk methods \cite{hong2021learning}, combining reinforcement learning with beam search techniques \cite{chatterjee2021weakly}, enhancing data quality by diversifying pseudo-samples \cite{lianggeneralizing} and enhancing recall through search space reduction \cite{liu2022comsearch}. Nevertheless, weakly supervised methods often overlook semantic alignment between problem statements and equations.

Recent advancements have leveraged LLM for data annotation in MWPs. \cite{raffel2023mathematical} introduced LLMs for assisting in MWP solving, while Gan et al. \cite{gan2023usa,SCICOT} proposed LLM-assisted data annotation techniques. Furthermore, Self-alignment \cite{Self-alignment} utilizes seed data to acquire high-quality dialogue data. This reveals the potential of LLMs in annotating high-quality data.

\appendix

\end{document}